\documentclass[10pt,twocolumn,letterpaper]{article}
\usepackage{cvpr}
\usepackage{times}
\usepackage{epsfig}
\usepackage{graphicx}
\usepackage{amsmath}
\usepackage{amssymb}
\usepackage{algorithm,algpseudocode,amsthm,mathrsfs,subcaption}
\usepackage{mathrsfs,enumitem}
\usepackage{multirow}
\usepackage{forloop}

\usepackage{xr}
\def\R{\mathbb{R}}

\def\Pbb{\mathbb{P}}

\graphicspath{{images/}{../Figs_val_googlenet_linf10_l2perts_PDF/}}

\usepackage[pagebackref=true,breaklinks=true,letterpaper=true,colorlinks,bookmarks=false]{hyperref}

\cvprfinalcopy 

\def\cvprPaperID{649} 

\ifcvprfinal\fi
\begin{document}

\title{Universal adversarial perturbations}

\author{Seyed-Mohsen Moosavi-Dezfooli\thanks{The first two authors contributed equally to this work.}\;\,\thanks{\'Ecole Polytechnique F\'ed\'erale de Lausanne, Switzerland} \\
{\tt\small seyed.moosavi@epfl.ch}
\and
Alhussein Fawzi\footnotemark[1]\;\,\footnotemark[2]\\
{\tt\small alhussein.fawzi@epfl.ch}
\and
\hspace{6cm}\\
{\hspace{6cm}}
\and
\hspace{-6.5cm}Omar Fawzi\thanks{ENS de Lyon, LIP, UMR 5668 ENS Lyon - CNRS - UCBL - INRIA, Universit\'e de Lyon, France}\\
{\hspace{-6.5cm}\tt\small omar.fawzi@ens-lyon.fr}
\and
\hspace{-1cm}Pascal Frossard\footnotemark[2] \\
{\hspace{-1cm}\tt\small pascal.frossard@epfl.ch}
}
\maketitle
\begin{abstract}
Given a state-of-the-art deep neural network classifier, we show the existence of a \underline{universal} (image-agnostic) and very small perturbation vector that causes natural images to be misclassified with high probability. We propose a systematic algorithm for computing universal perturbations, and show that state-of-the-art deep neural networks are highly vulnerable to such perturbations, albeit being quasi-imperceptible to the human eye.
We further empirically analyze these universal perturbations and show, in particular, that they generalize very well across neural networks. The surprising existence of universal perturbations reveals important geometric \textit{correlations} among the high-dimensional decision boundary of classifiers. It further outlines potential security breaches with the existence of single directions in the input space that adversaries can possibly exploit to break a classifier on most natural images.\footnote{To encourage reproducible research, the code is available at \href{https://github.com/LTS4/universal}{gitHub}. Furthermore, a video demonstrating the effect of universal perturbations on a smartphone can be found \href{https://youtu.be/jhOu5yhe0rc}{here}.}
\end{abstract}


\section{Introduction}

\begin{figure}[ht!]
\center
\includegraphics[scale=0.35]{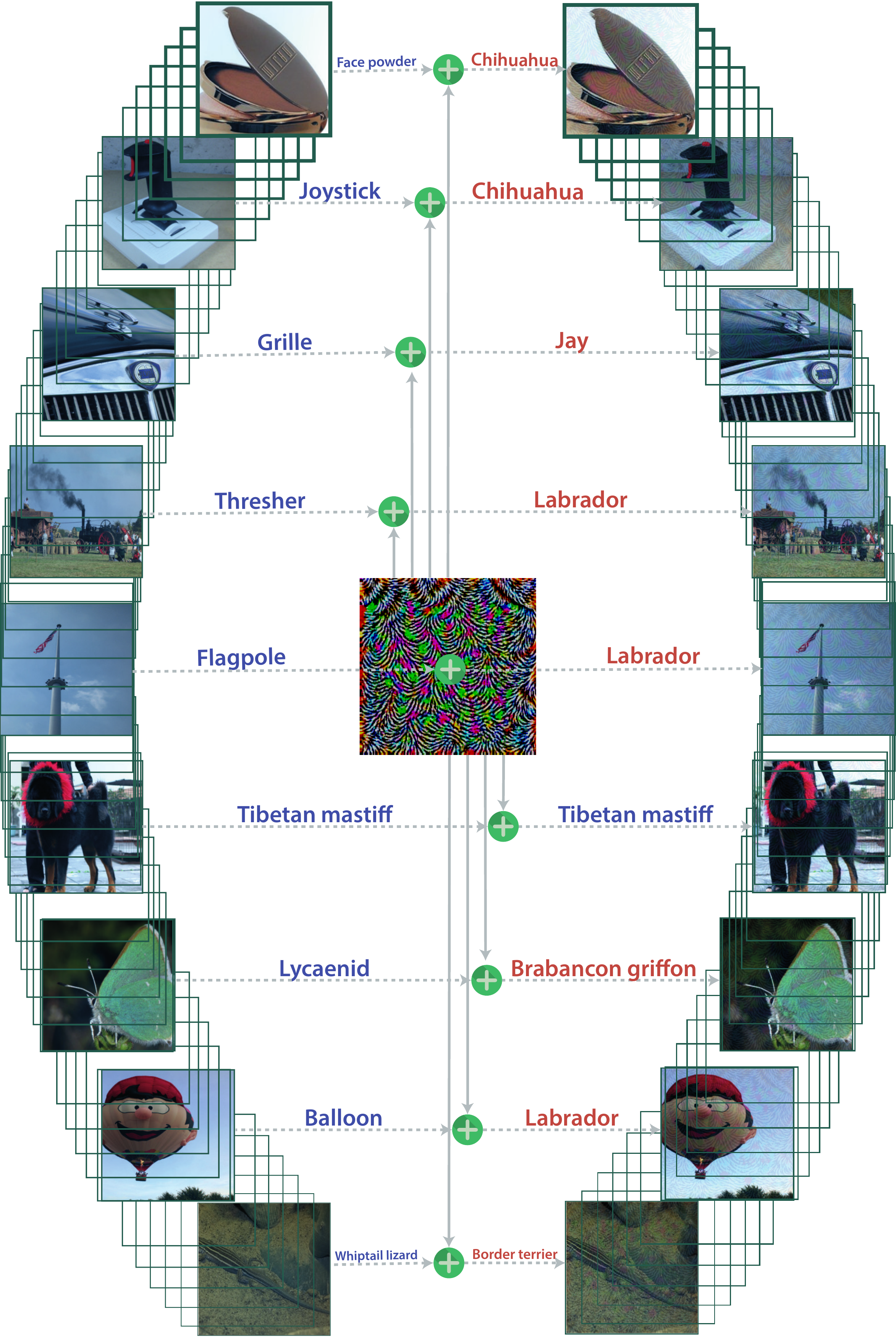}
\caption{\label{fig:main_fig} When added to a natural image, a universal perturbation image causes the image to be misclassified by the deep neural network with high probability. \textit{Left images:} Original natural images. The labels are shown on top of each arrow. \textit{Central image:} Universal perturbation. \textit{Right images:} Perturbed images. The estimated labels of the perturbed images are shown on top of each arrow.}
\end{figure}

Can we find a \textit{single} small image perturbation that fools a state-of-the-art deep neural network classifier on all natural images? We show in this paper the existence of such quasi-imperceptible \textit{universal} perturbation vectors that lead to misclassify natural images with high probability. Specifically, by adding such a \textit{quasi-imperceptible} perturbation to natural images, the label estimated by the deep neural network is changed with high probability (see Fig. \ref{fig:main_fig}). Such perturbations are dubbed \textit{universal}, as they are image-agnostic. The existence of these perturbations is problematic when the classifier is deployed in real-world (and possibly hostile) environments, as they can be exploited by adversaries to break the classifier. Indeed, the perturbation process involves the mere addition of one very small perturbation to all natural images, and can be relatively straightforward to implement by adversaries in real-world environments, while being relatively difficult to detect as such perturbations are very small and thus do not significantly affect data distributions. 
The surprising existence of universal perturbations further reveals new insights on the topology of the decision boundaries of deep neural networks. 
We summarize the main contributions of this paper as follows:
\begin{itemize}
\item We show the existence of universal image-agnostic perturbations for state-of-the-art deep neural networks.
\item We propose an algorithm for finding such perturbations. The algorithm seeks a universal perturbation for a set of training points, and proceeds by aggregating atomic perturbation vectors that send successive datapoints to the decision boundary of the classifier. 
\item We show that universal perturbations have a remarkable generalization property, as perturbations computed for a rather small set of training points fool new images with high probability.
\item We show that such perturbations are not only universal across images, but also generalize well across deep neural networks. Such perturbations are therefore \textit{doubly} universal, both with respect to the data and the network architectures.
\item We explain and analyze the high vulnerability of deep neural networks to universal perturbations by examining the geometric correlation between different parts of the decision boundary. 
\end{itemize}

The robustness of image classifiers to structured and unstructured perturbations have recently attracted a lot of attention \cite{szegedy2013intriguing, sabour2016adversarial, tabacof2015exploring, fawzi2015a, nips2016_ours, nguyen2015, rodner2016, Rozsa_2016_CVPR_Workshops}. Despite the impressive performance of deep neural network architectures on challenging visual classification benchmarks \cite{he2015deep, cv2, taigman2014deepface, le2011learning}, these classifiers were shown to be highly vulnerable to perturbations. In \cite{szegedy2013intriguing}, such networks are shown to be
 unstable to very small and often imperceptible additive \textit{adversarial} perturbations. Such carefully crafted perturbations are either estimated by solving an optimization problem \cite{szegedy2013intriguing, moosavi2015deepfool, bastani2016measuring} or through one step of gradient ascent \cite{goodfellow2014}, and result in a perturbation that fools a specific data point. A fundamental property of these adversarial perturbations is their intrinsic dependence on datapoints: the perturbations are specifically crafted for each data point independently. As a result, the computation of an adversarial perturbation for a new data point requires solving a data-dependent optimization problem from scratch, which uses the full knowledge of the classification model. 
This is different from the universal perturbation considered in this paper, as we seek a single perturbation vector that fools the network on most natural images. Perturbing a new datapoint then only involves the mere addition of the universal perturbation to the image (and does not require solving an optimization problem/gradient computation). Finally, we emphasize that our notion of universal perturbation differs from the generalization of adversarial perturbations studied in \cite{szegedy2013intriguing}, where perturbations computed on the MNIST task were shown to generalize well across different models. Instead, we examine the existence of universal perturbations that are common to most data points belonging to the data distribution.

\section{Universal perturbations}

We formalize in this section the notion of universal perturbations, and propose a method for estimating such perturbations. Let $\mu$ denote a distribution of images in $\R^d$, and $\hat{k}$ define a classification function that outputs for each image $x \in \R^d$ an estimated label $\hat{k}(x)$. The main focus of this paper is to seek perturbation vectors $v \in \R^d$ that fool the classifier $\hat{k}$ on \textit{almost all} datapoints sampled from $\mu$. That is, we seek a vector $v$ such that
$$ \hat{k} (x+v) \neq \hat{k} (x) \text{ for ``most'' } x \sim \mu.$$
We coin such a perturbation \textit{universal}, as it represents a fixed image-agnostic perturbation that causes label change for most images sampled from the data distribution $\mu$. We focus here on the case where the distribution $\mu$ represents the set of natural images, hence containing a huge amount of variability. In that context, we
examine the existence of small universal perturbations (in terms of the $\ell_p$ norm with $p \in [1, \infty)$) that misclassify most images. The goal is therefore to find $v$ that satisfies the following two constraints: 
\begin{enumerate}[topsep=3pt,itemsep=0ex,partopsep=1ex,parsep=1ex]
\item $\| v \|_p \leq \xi,$
\item $\underset{x \sim \mu}{\Pbb} \left( \hat{k} (x+v) \neq \hat{k} (x) \right) \geq 1 - \delta.$
\end{enumerate}
%
The parameter $\xi$ controls the magnitude of the perturbation vector $v$, and $\delta$ quantifies the desired fooling rate for all images sampled from the distribution $\mu$.




\vspace{2mm}

\begin{figure}[t]
\centering
\includegraphics[scale=0.5]{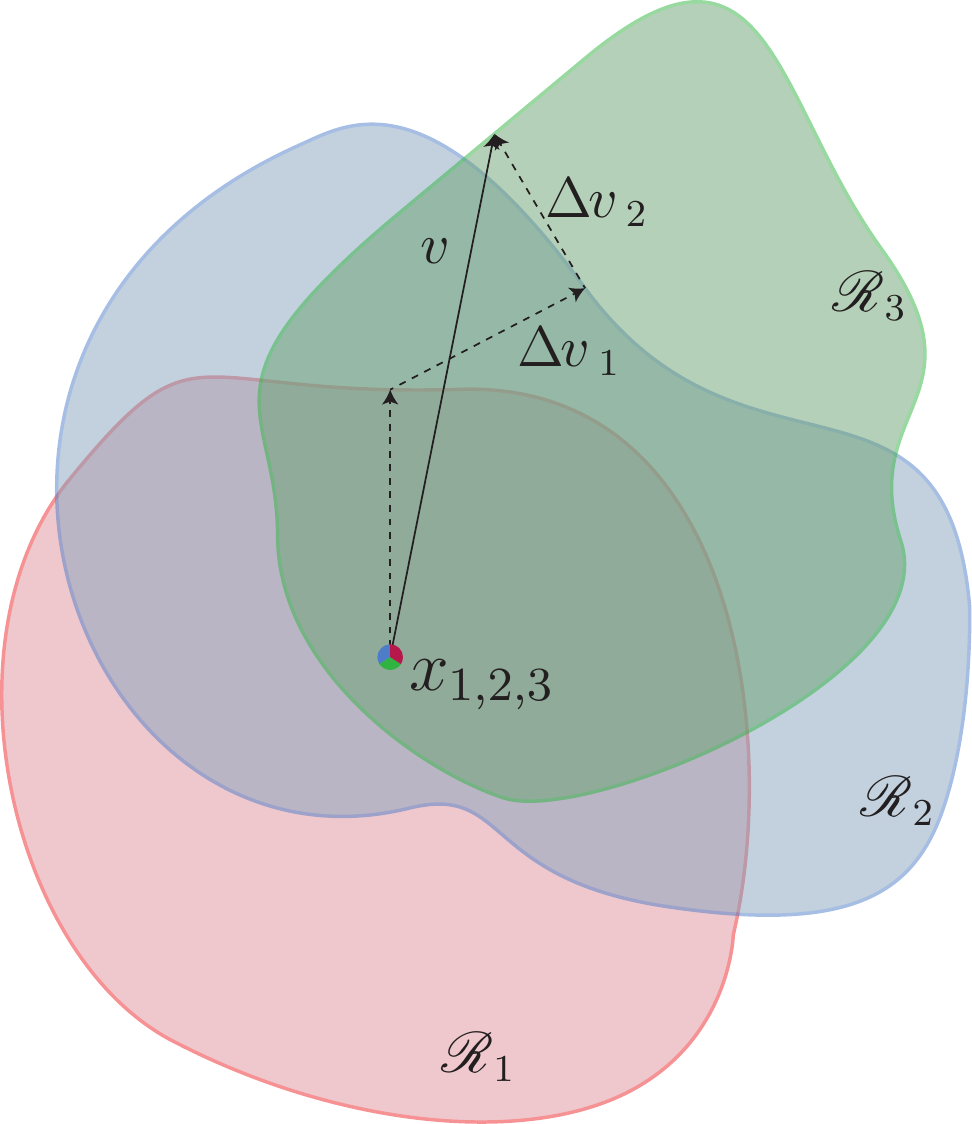}
\caption{\label{fig:schematic_representation_algo} Schematic representation of the proposed algorithm used to compute universal perturbations. In this illustration, data points $x_1, x_2$ and $x_3$ are super-imposed, and the classification regions $\mathscr{R}_i$ (i.e., regions of constant estimated label) are shown in different colors. Our algorithm proceeds by aggregating sequentially the minimal perturbations sending the current perturbed points $x_i+v$ outside of the corresponding classification region $\mathscr{R}_i$.}
\end{figure}

\textbf{Algorithm.} Let $X = \{ x_1, \dots, x_m \}$ be a set of images sampled from the distribution $\mu$. Our proposed algorithm seeks a universal perturbation $v$, such that $\| v \|_p \leq \xi$, while fooling most data points in $X$. 
The algorithm proceeds iteratively over the data points in $X$ and gradually builds the universal perturbation, as illustrated in Fig. \ref{fig:schematic_representation_algo}. At each iteration, the minimal perturbation $\Delta v_i$ that sends the current perturbed point, $x_i + v$, to the decision boundary of the classifier is computed, and aggregated to the current instance of the universal perturbation. In more details, provided the current universal perturbation $v$ does not fool data point $x_i$, we seek the extra perturbation $\Delta v_i$ with minimal norm that allows to fool data point $x_i$ by solving the following optimization problem:
\begin{align}
\label{eq:project_on_boundary}
\Delta v_i & \gets \arg\min_{r} \| r \|_2 \text{ s.t. } \hat{k} (x_i + v + r) \neq \hat{k} (x_i).
\end{align}
To ensure that the constraint $\| v \|_p \leq \xi$ is satisfied, the updated universal perturbation is further projected on the $\ell_p$ ball of radius $\xi$ and centered at $0$. That is, let $\mathcal{P}_{p, \xi}$ be the projection operator defined as follows:
\begin{align*}
\mathcal{P}_{p, \xi} (v) = \arg\min_{v'} \| v - v' \|_2 \text{ subject to } \| v' \|_p \leq \xi.
\end{align*}
Then, our update rule is given by $v \gets \mathcal{P}_{p, \xi} (v + \Delta v_i)$.
Several passes on the data set $X$ are performed to improve the quality of the universal perturbation. The algorithm is terminated when the empirical ``fooling rate'' on the perturbed data set $X_v  := \{x_1 + v, \dots, x_m + v\}$ exceeds the target threshold $1-\delta$. That is, we stop the algorithm whenever
\begin{align*}
\text{Err}(X_v) := \frac{1}{m} \sum_{i=1}^m 1_{\hat{k} (x_i+v) \neq \hat{k}(x_i)} \geq 1-\delta.
\end{align*}
The detailed algorithm is provided in Algorithm \ref{alg:finding_universal_perturbations}. Interestingly, in practice, the number of data points $m$ in $X$ need not be large to compute a universal perturbation that is valid for the whole distribution $\mu$. In particular, we can set $m$ to be much smaller than the number of training points (see Section \ref{sec:experiments}).


\begin{algorithm}[t]
\caption{Computation of universal perturbations.}
\begin{algorithmic}[1]
\State \textbf{input:} Data points $X$, classifier $\hat{k}$, desired $\ell_p$ norm of the perturbation $\xi$, desired accuracy on perturbed samples $\delta$.
\State \textbf{output:} Universal perturbation vector $v$.
\State Initialize $v \gets 0$.
\While{$\text{Err} (X_v) \leq 1-\delta$}
\For{each datapoint $x_i \in X$}
\If{$\hat{k}(x_i+v) = \hat{k} (x_i)$}
\State Compute the \textit{minimal} perturbation that sends $x_i+v$ to the decision boundary:
\begin{align*}
\quad \quad \Delta v_i \gets \arg\min_{r} \| r \|_2 \text{ s.t. } \hat{k} (x_i + v + r) \neq \hat{k} (x_i).
\end{align*}
\State Update the perturbation: 
$$v \gets \mathcal{P}_{p, \xi} (v+\Delta v_i).$$
\EndIf
\EndFor
\EndWhile
\end{algorithmic}
\label{alg:finding_universal_perturbations}
\end{algorithm}

The proposed algorithm involves solving at most $m$ instances of the optimization problem in Eq. (\ref{eq:project_on_boundary}) for each pass. While this optimization problem is not convex when $\hat{k}$ is a standard classifier (e.g., a deep neural network), several efficient approximate methods have been devised for solving this problem \cite{szegedy2013intriguing, moosavi2015deepfool, huang2015learning}. We use in the following the approach in \cite{moosavi2015deepfool} for its efficency. It should further be noticed that the objective of Algorithm \ref{alg:finding_universal_perturbations} is \textit{not} to find the smallest universal perturbation that fools most data points sampled from the distribution, but rather to find one such perturbation with sufficiently small norm. In particular, different random shufflings of the set $X$ naturally lead to a diverse set of universal perturbations $v$ satisfying the required constraints. The proposed algorithm can therefore be leveraged to generate multiple universal perturbations for a deep neural network (see next section for visual examples).




\section{Universal perturbations for deep nets}
\label{sec:experiments}
\begin{table*}
\centering
\begin{tabular}{|l|l|c|c|c|c|c|c|}
\hline
& & CaffeNet \cite{jia2014} & VGG-F \cite{chatfield2014} & VGG-16 \cite{simonyan2014very} & VGG-19 \cite{simonyan2014very} & GoogLeNet \cite{szegedy2015} & ResNet-152 \cite{he2015deep} \\ \hline
\multirow{ 2}{*}{$\ell_{2}$} & $X$ & 85.4\% & 85.9\% & 90.7\%  & 86.9\% & 82.9\%  & 89.7\% \\
& Val. & 85.6 & 87.0\% & 90.3\% & 84.5\% & 82.0\%  & 88.5\% \\ \hline
\multirow{ 2}{*}{$\ell_{\infty}$} & $X$ & 93.1\% & 93.8\% & 78.5\% & 77.8\% & 80.8\% & 85.4\%\\
& Val. & 93.3\% & 93.7\% & 78.3\% & 77.8\% & 78.9\%  & 84.0\% \\ \hline
\end{tabular}
\caption{\label{tab:fooling_ratios} Fooling ratios on the set $X$, and the validation set.}
\end{table*}

We now analyze the robustness of state-of-the-art deep neural network classifiers to universal perturbations using Algorithm \ref{alg:finding_universal_perturbations}.

\begin{figure*}[ht]
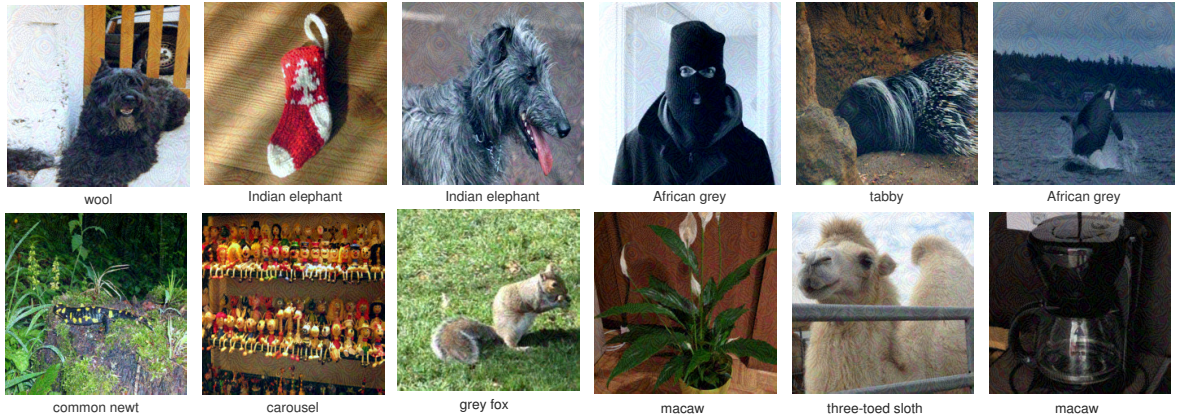

\centering
\newcounter{perturbed}
\forloop{perturbed}{1}{\value{perturbed} < 9}{
\begin{subfigure}[t]{0.14\textwidth}
\includegraphics[width=\textwidth]{image_perturbedonly_\arabic{perturbed}.pdf}
\end{subfigure}
}
\hspace{-2mm}
\newcounter{natperturbed}
\forloop{natperturbed}{1}{\value{natperturbed} < 5}{
\begin{subfigure}[t]{0.14\textwidth}
\includegraphics[width=\textwidth]{natimg\arabic{natperturbed}_cropped_googlenet_pert.pdf}
\end{subfigure}
}
\caption{\label{fig:perturbed_images} Examples of perturbed images and their corresponding labels. The first $8$ images belong to the ILSVRC 2012 validation set, and the last $4$ are images taken by a mobile phone camera. See supp. material for the original images.}
\end{figure*}
\begin{figure*}[ht]
\centering
\begin{subfigure}[t]{0.23\textwidth}
\includegraphics[width=\textwidth]{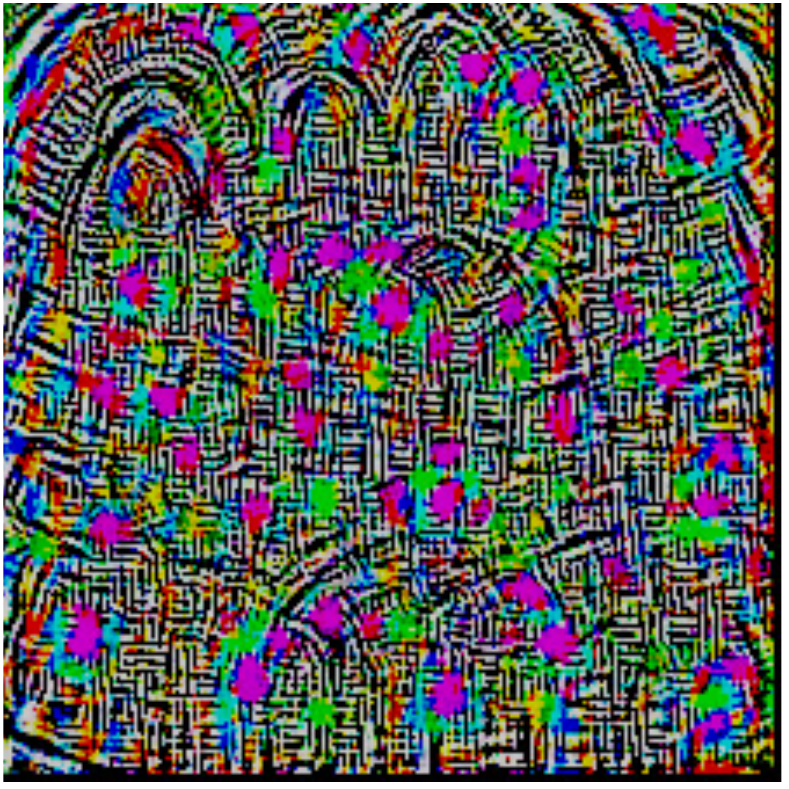}
\caption{\label{lab:caffenet_pert} CaffeNet}
\end{subfigure}
~
\begin{subfigure}[t]{0.23\textwidth}
\includegraphics[width=\textwidth]{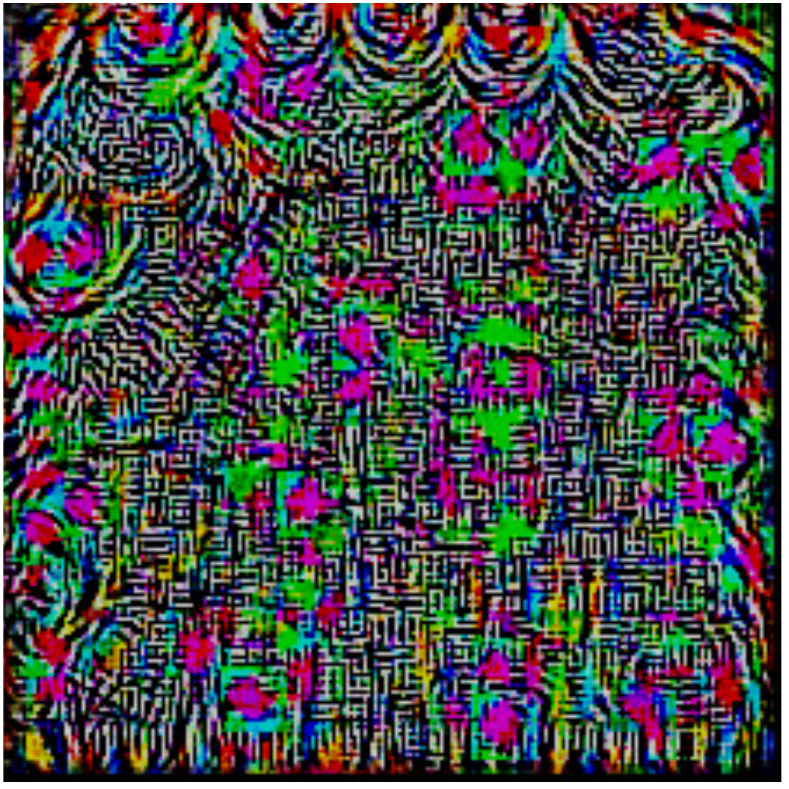}
\caption{\label{lab:vgg_f_pert} VGG-F}
\end{subfigure}
~
\begin{subfigure}[t]{0.23\textwidth}
\includegraphics[width=\textwidth]{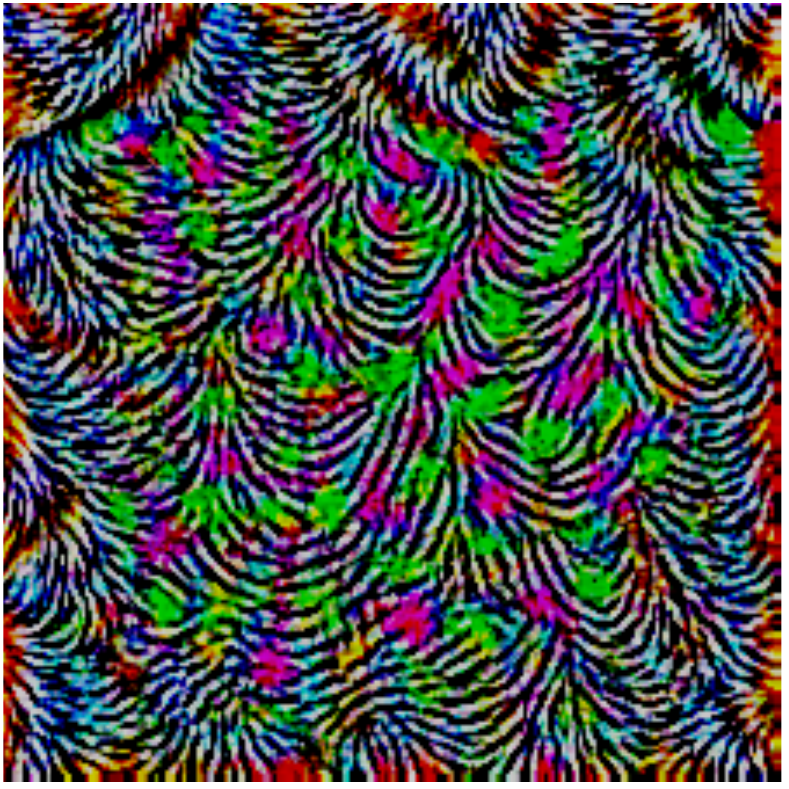}
\caption{\label{lab:vgg_16_pert} VGG-16}
\end{subfigure}
\\
\begin{subfigure}[t]{0.23\textwidth}
\includegraphics[width=\textwidth]{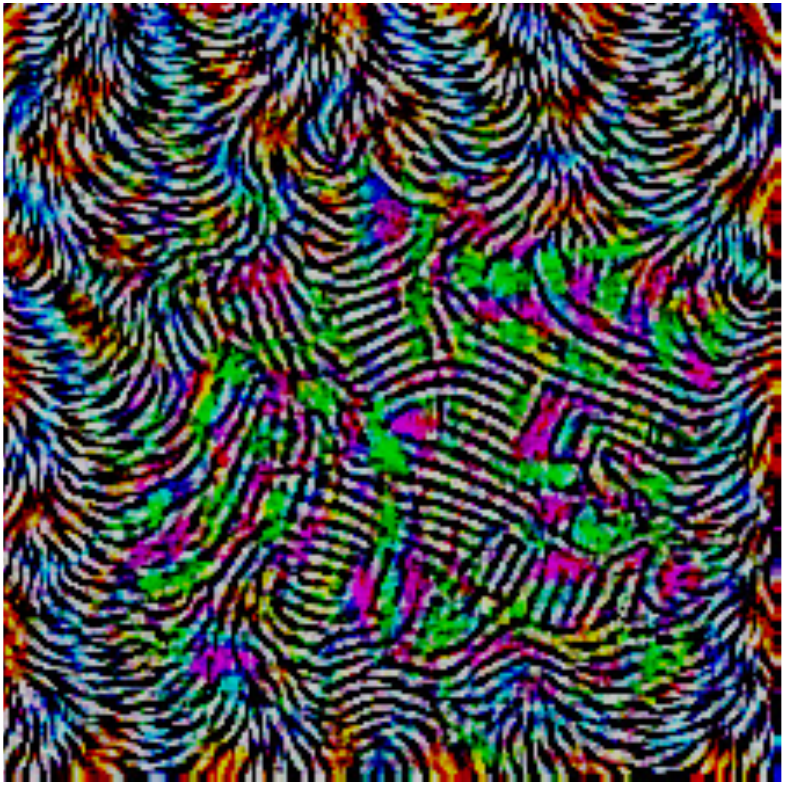}
\caption{\label{lab:vgg_19_pert} VGG-19}
\end{subfigure}
~
\begin{subfigure}[t]{0.23\textwidth}
\includegraphics[width=\textwidth]{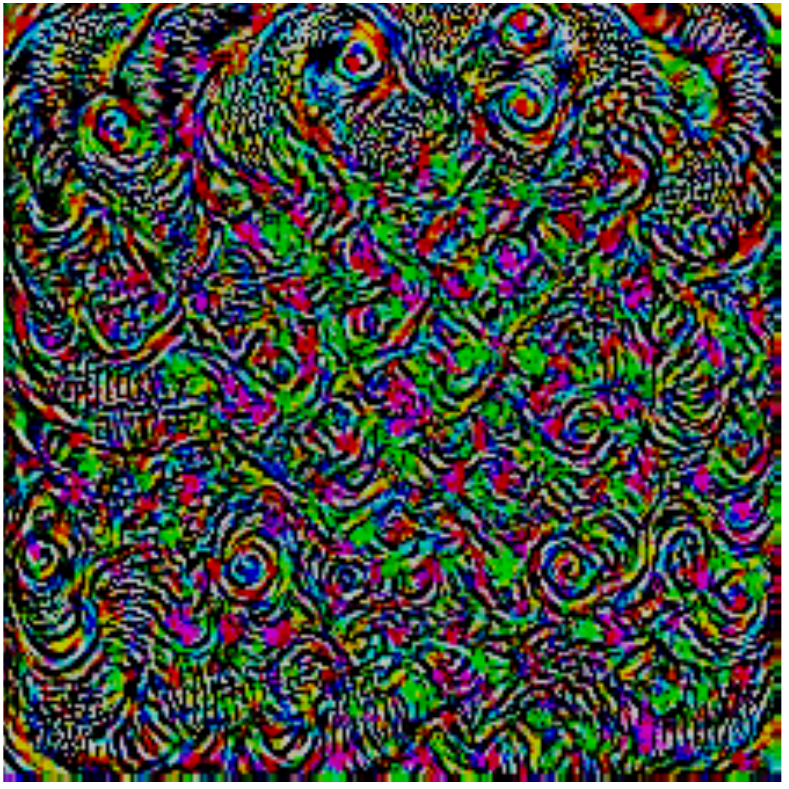}
\caption{\label{lab:googlenet} GoogLeNet}
\end{subfigure}
~
\begin{subfigure}[t]{0.23\textwidth}
\includegraphics[width=\textwidth]{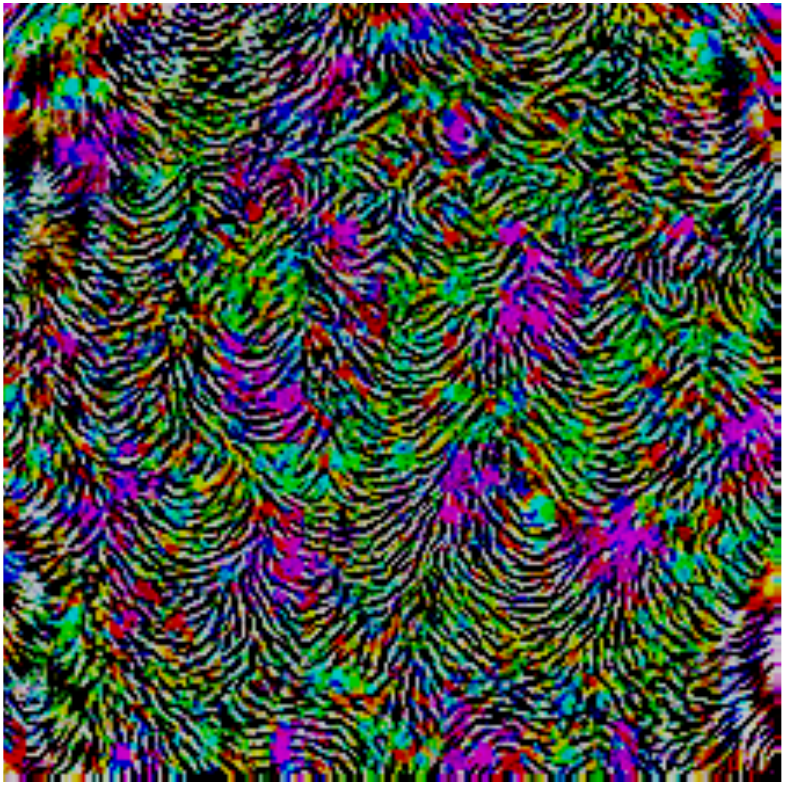}
\caption{\label{lab:resnet} ResNet-152}
\end{subfigure}
\caption{\label{fig:universal_perturbation_vectors} Universal perturbations computed for different deep neural network architectures. Images generated with $p=\infty$, $\xi = 10$. The pixel values are scaled for visibility.}
\end{figure*}
\begin{figure*}[ht]
\centering
\begin{subfigure}[t]{0.19\textwidth}
\includegraphics[width=\textwidth]{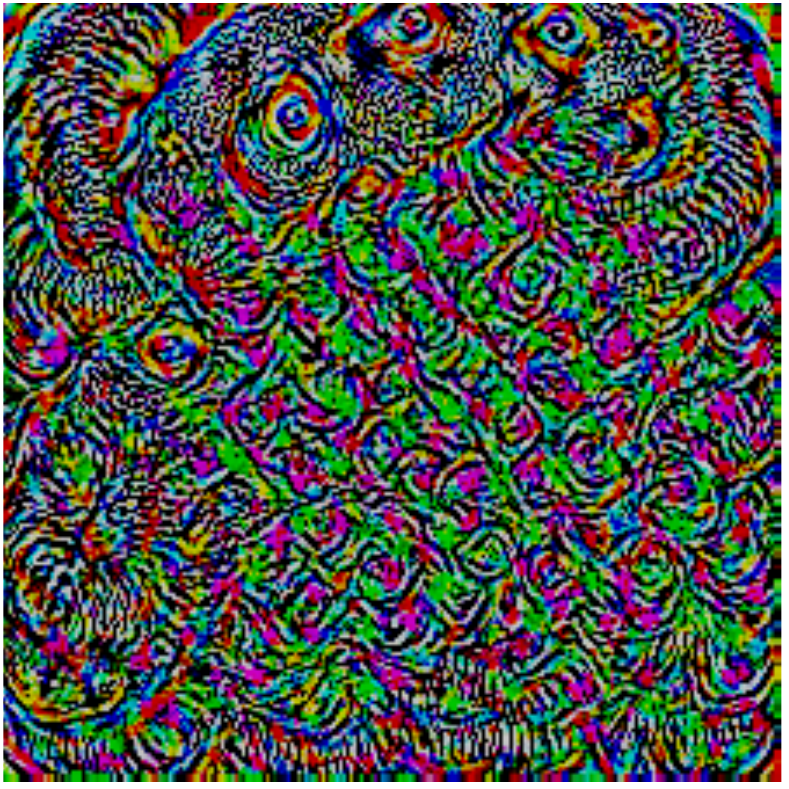}
\end{subfigure}
\begin{subfigure}[t]{0.19\textwidth}
\includegraphics[width=\textwidth]{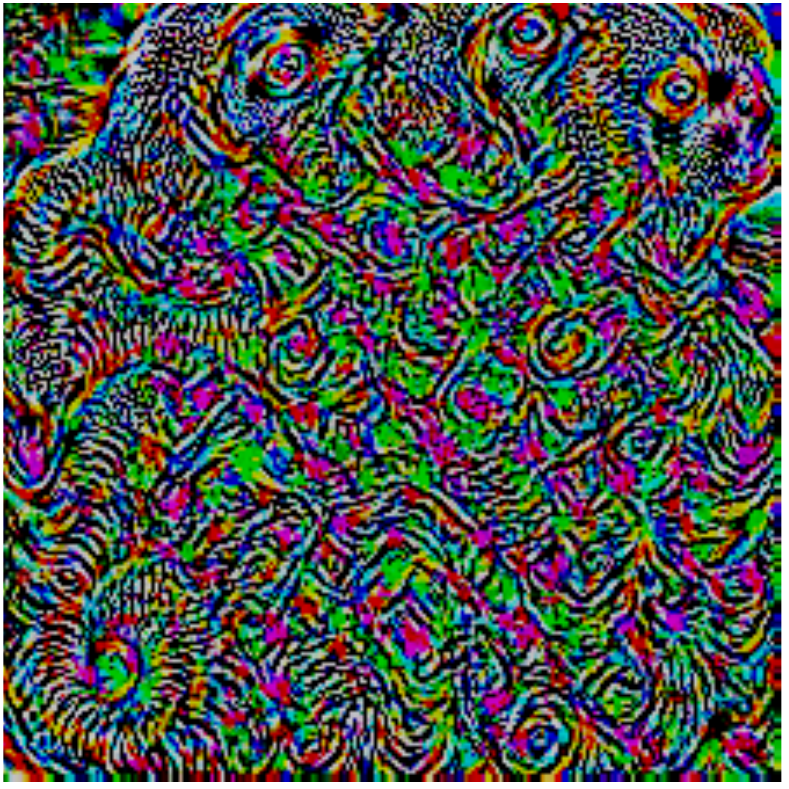}
\end{subfigure}
\begin{subfigure}[t]{0.19\textwidth}
\includegraphics[width=\textwidth]{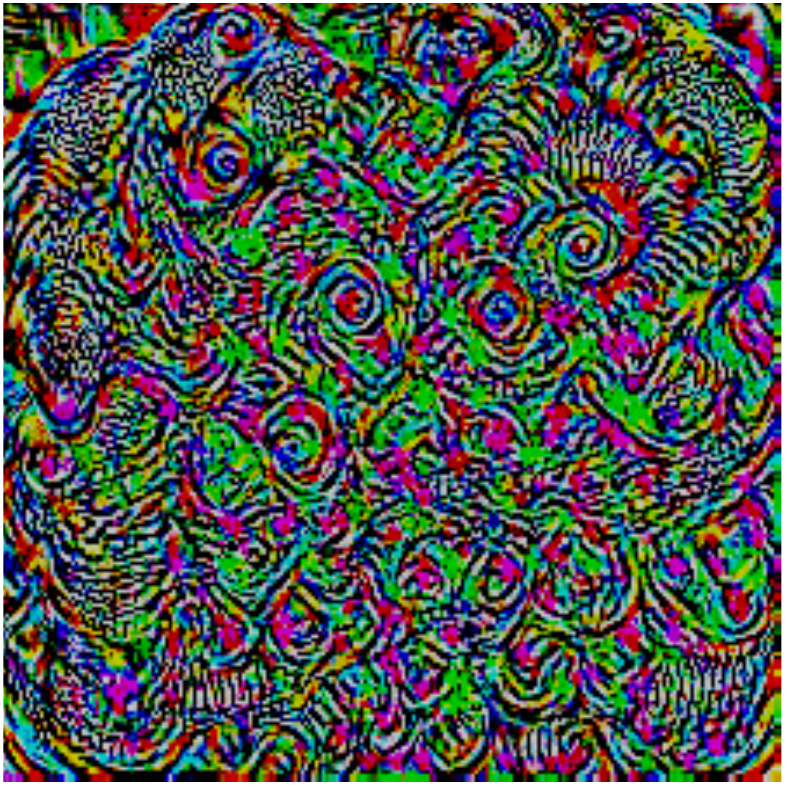}
\end{subfigure}
\begin{subfigure}[t]{0.19\textwidth}
\includegraphics[width=\textwidth]{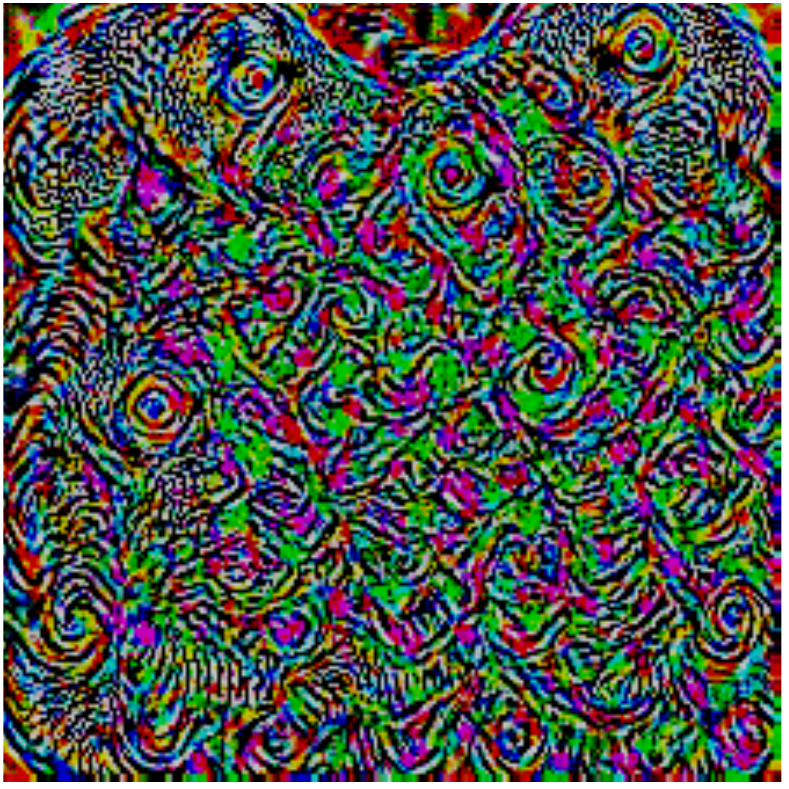}
\end{subfigure}
\begin{subfigure}[t]{0.19\textwidth}
\includegraphics[width=\textwidth]{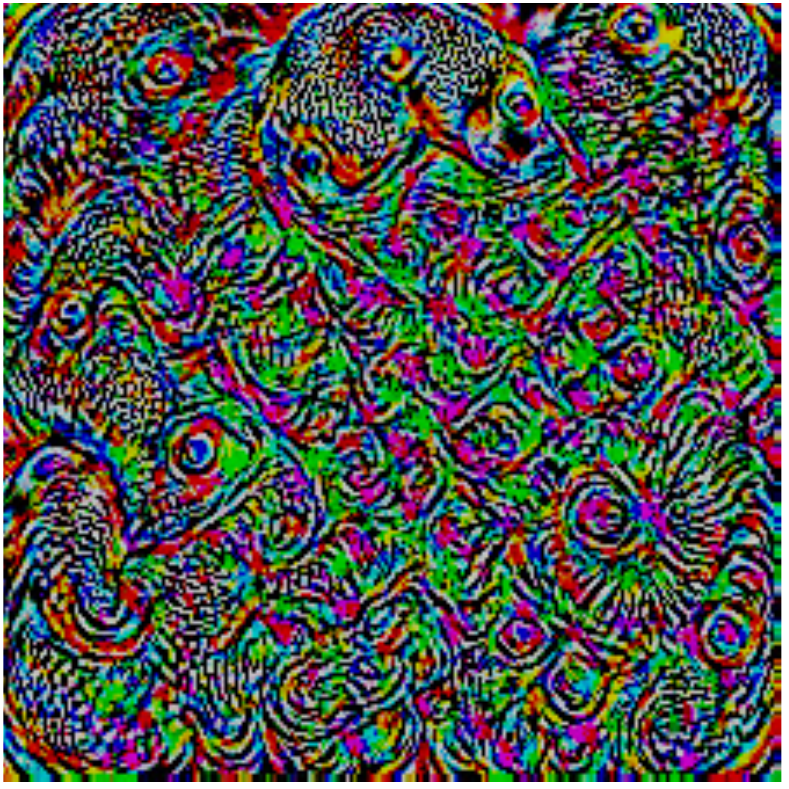}
\end{subfigure}
\caption{\label{fig:diversity_univ_perts} Diversity of universal perturbations for the GoogLeNet architecture. The five perturbations are generated using different random shufflings of the set $X$. Note that the normalized inner products for any pair of universal perturbations does not exceed $0.1$, which highlights the diversity of such perturbations.}
\end{figure*}

In a first experiment, we assess the estimated universal perturbations for different recent deep neural networks on the ILSVRC 2012 \cite{russakovsky2015imagenet} validation set (50,000 images), and report the \textit{fooling ratio}, that is the proportion of images that change labels when perturbed by our universal perturbation. Results are reported for $p=2$ and $p=\infty$, where we respectively set $\xi = 2000$ and $\xi = 10$. These numerical values were chosen in order to obtain a perturbation whose norm is significantly smaller than the image norms, such that the perturbation is quasi-imperceptible when added to natural images\footnote{For comparison, the average $\ell_2$ and $\ell_{\infty}$ norm of an image in the validation set is respectively $\approx 5 \times 10^4$ and $\approx 250$.}. Results are listed in Table \ref{tab:fooling_ratios}. 
Each result is reported on the set $X$, which is used to compute the perturbation, as well as on the validation set (that is \textit{not} used in the process of the computation of the universal perturbation). Observe that for all networks, the universal perturbation achieves very high fooling rates on the validation set. Specifically, the universal perturbations computed for CaffeNet and VGG-F fool more than $90\%$ of the validation set (for $p=\infty$). In other words, for any natural image in the validation set, the mere addition of our universal perturbation fools the classifier more than $9$ times out of $10$. This result is moreover not specific to such architectures, as we can also find universal perturbations that cause VGG, GoogLeNet and ResNet classifiers to be fooled on natural images with probability edging $80\%$. 
These results have an element of surprise, as they show the
existence of \textit{single} universal perturbation vectors that cause natural images to be misclassified with high probability, albeit being quasi-imperceptible to humans. To verify this latter claim, we show visual examples of perturbed images in Fig. \ref{fig:perturbed_images}, where the GoogLeNet architecture is used. These images are either taken from the ILSVRC 2012 validation set, or captured using a mobile phone camera. 
Observe that in most cases, the universal perturbation is \textit{quasi-imperceptible}, yet this powerful image-agnostic perturbation is able to misclassify any image with high probability for state-of-the-art classifiers. We refer to the supp. material for the original (unperturbed) images, as well as their ground truth labels. We also refer to the video in the supplementary material for real-world examples on a smartphone. We visualize the universal perturbations corresponding to different networks in Fig. \ref{fig:universal_perturbation_vectors}. It should be noted that such universal perturbations are not unique, as many different universal perturbations (all satisfying the two required constraints) can be generated for the same network. In Fig. \ref{fig:diversity_univ_perts}, we visualize five different universal perturbations obtained by using different random shufflings in $X$. Observe that such universal perturbations are different, although they exhibit a similar pattern. This is moreover confirmed by computing the normalized inner products between two pairs of perturbation images, as the normalized inner products do not exceed $0.1$, which shows that one can find diverse universal perturbations.

\begin{figure}[ht]
\centering
\includegraphics[scale=0.40]{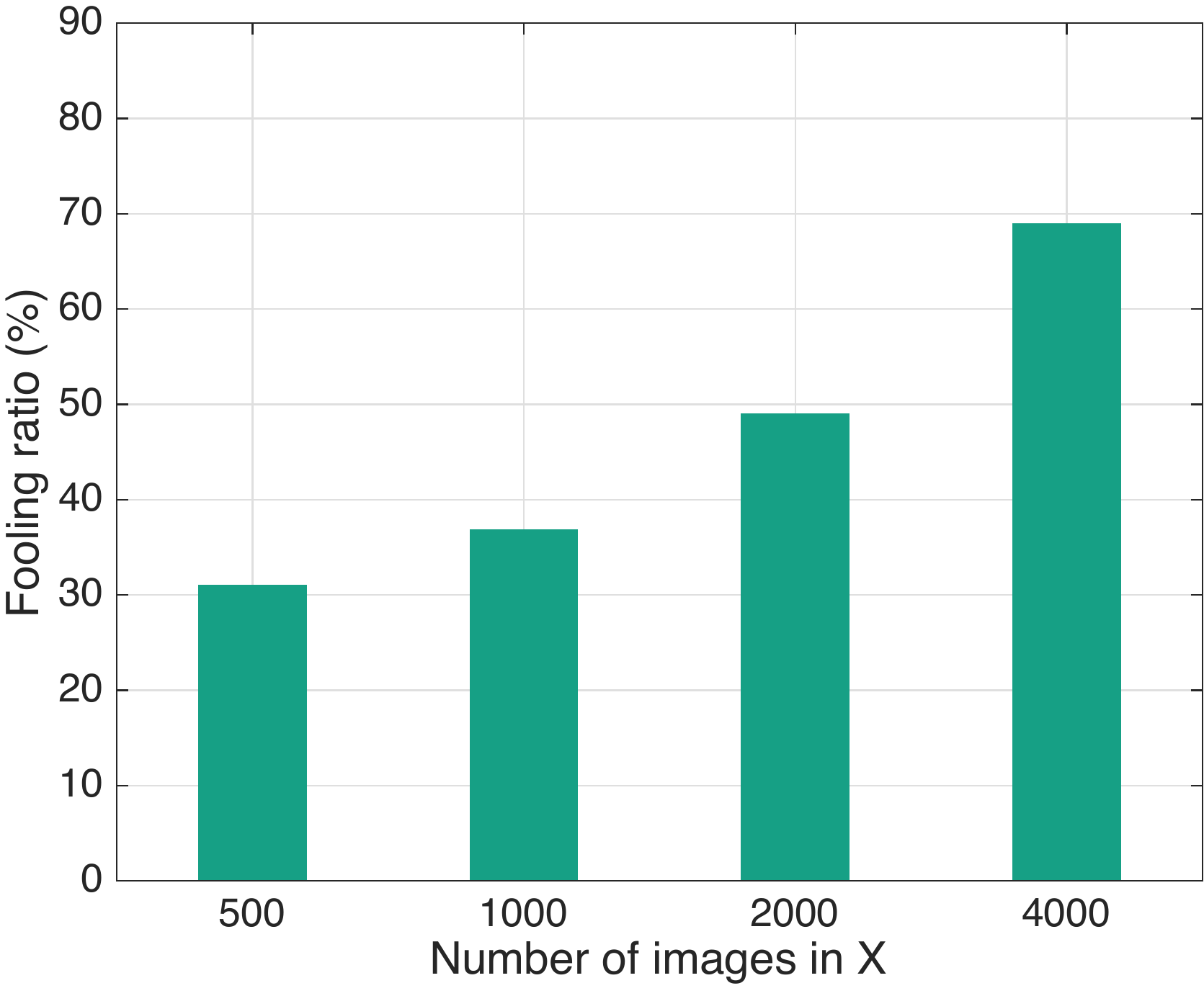}
\caption{\label{fig:histogram_linf_numDatapointsX_val} Fooling ratio on the validation set versus the size of $X$. Note that even when the universal perturbation is computed on a very small set $X$ (compared to training and validation sets), the fooling ratio on validation set is large.}
\end{figure}

While the above universal perturbations are computed for a set $X$ of 10,000 images from the training set (i.e., in average $10$ images per class), we now examine the influence of the size of $X$ on the quality of the universal perturbation. We show in Fig. \ref{fig:histogram_linf_numDatapointsX_val} the fooling rates obtained on the validation set for different sizes of $X$ for GoogLeNet. Note for example that with a set $X$ containing only $500$ images, we can fool more than $30\%$ of the images on the validation set. This result is significant when compared to the number of classes in ImageNet ($1000$), as it shows that we can fool a large set of unseen images, even when using a set $X$ containing less than one image per class! The universal perturbations computed using Algorithm \ref{alg:finding_universal_perturbations} have therefore a remarkable generalization power over unseen data points, and can be computed on a very small set of training images. 

\textbf{Cross-model universality.} While the computed perturbations are universal across unseen data points, we now examine their \textit{cross-model} universality. That is, we study to which extent universal perturbations computed for a specific architecture (e.g., VGG-19) are also valid for another architecture (e.g., GoogLeNet). Table \ref{tab:generalization_across_networks} displays a matrix summarizing the universality of such perturbations across six different architectures. 
For each architecture, we compute a universal perturbation and report the fooling ratios on all other architectures; we report these in the rows of Table \ref{tab:generalization_across_networks}.
Observe that, for some architectures, the universal perturbations generalize very well across other architectures. For example, universal perturbations computed for the VGG-19 network have a fooling ratio above $53\%$ for all other tested architectures. This result shows that our universal perturbations are, to some extent, \textit{doubly-universal} as they generalize well across data points \textit{and} very different architectures. It should be noted that, in \cite{szegedy2013intriguing}, adversarial perturbations were previously shown to generalize well, to some extent, across different neural networks on the MNIST problem. Our results are however different, as we show the generalizability of universal perturbations across different architectures on the ImageNet data set. This result shows that such perturbations are of practical relevance, as they generalize well across data points and architectures. In particular, in order to fool a new image on an unknown neural network, a simple addition of a universal perturbation computed on the VGG-19 architecture is likely to misclassify the data point.
\begin{table*}
\centering
\begin{tabular}{l|l|l|l|l|l|l|}
\cline{2-7}
                                & VGG-F & CaffeNet & GoogLeNet & VGG-16 & VGG-19 & ResNet-152 \\ \hline
\multicolumn{1}{|l|}{VGG-F}     & \textbf{93.7\%}& 71.8\%   & 48.4\%    & 42.1\% & 42.1\% & 47.4 \%\\ \hline
\multicolumn{1}{|l|}{CaffeNet}  &  74.0\%  & \textbf{93.3\%}     & 47.7\%      & 39.9\%     & 39.9\% & 48.0\%   \\ \hline
\multicolumn{1}{|l|}{GoogLeNet} &46.2\% &43.8\%    & \textbf{78.9\%}          &  39.2\%      & 39.8\% & 45.5\%     \\ \hline
\multicolumn{1}{|l|}{VGG-16}    &   63.4\%    & 55.8\%   & 56.5\%   & \textbf{78.3\%}       & 73.1\% &  63.4\%     \\ \hline
\multicolumn{1}{|l|}{VGG-19}    & 64.0\%& 57.2\%   & 53.6\%    & 73.5\% & \textbf{77.8\%} & 58.0\% \\ \hline
\multicolumn{1}{|l|}{ResNet-152}    & 46.3\%& 46.3\%   & 50.5\%    & 47.0\% & 45.5\% & \textbf{84.0\%} \\ \hline
\end{tabular}
\caption{Generalizability of the universal perturbations across different networks. The percentages indicate the fooling rates. The rows indicate the architecture for which the universal perturbations is computed, and the columns indicate the architecture for which the fooling rate is reported.}
\label{tab:generalization_across_networks}
\end{table*}

 \vspace{2mm}

\begin{figure*}
\centering
\begin{subfigure}[t]{0.35\textwidth}
\includegraphics[width=\textwidth]{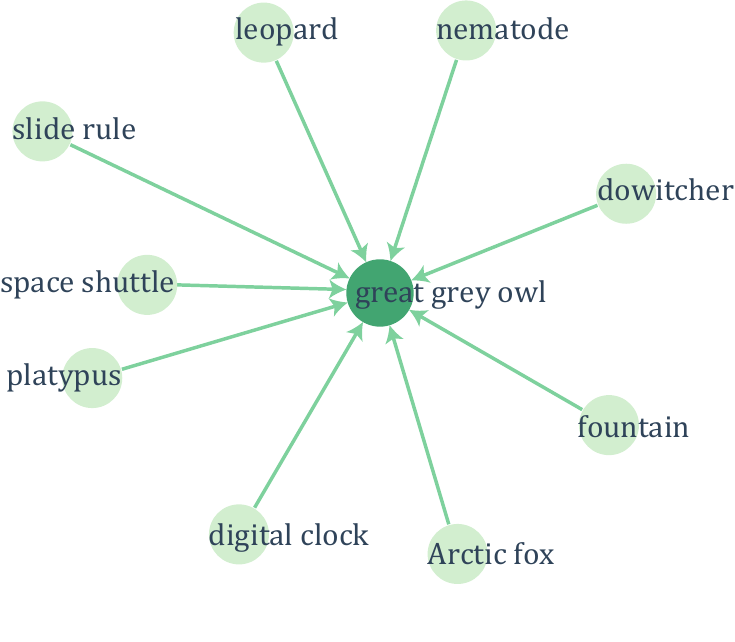}
\end{subfigure}
\begin{subfigure}[t]{0.35\textwidth}
\includegraphics[width=\textwidth]{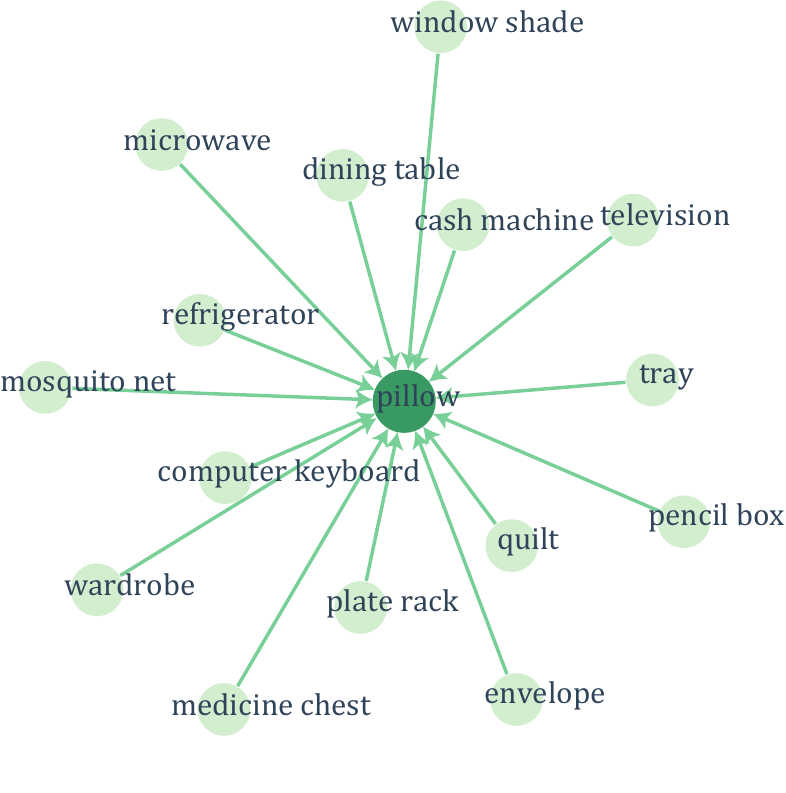}
\end{subfigure}
\caption{\label{fig:connected_components_googLeNet} Two connected components of the graph $G = (V,E)$, where the vertices are the set of labels, and directed edges $i \rightarrow j$ indicate that most images of class $i$ are fooled into class $j$.}
\end{figure*}

\textbf{Visualization of the effect of universal perturbations.} To gain insights on the effect of universal perturbations on natural images, we now visualize the distribution of labels on the ImageNet validation set. Specifically, we build a directed graph $G = (V, E)$, whose vertices denote the labels, and directed edges $e = (i \rightarrow j)$ indicate that the majority of images of class $i$ are fooled into label $j$ when applying the universal perturbation. The existence of edges $i \rightarrow j$ therefore suggests that the preferred fooling label for images of class $i$ is $j$.
We construct this graph for GoogLeNet, and visualize the full graph in the supp. material for space constraints. The visualization of this graph shows a very peculiar topology. In particular, the graph is a union of disjoint components, where all edges in one component mostly connect to one target label. See Fig. \ref{fig:connected_components_googLeNet} for an illustration of two connected components. This visualization clearly shows the existence of several \textit{dominant labels}, and that universal perturbations mostly make natural images classified with such labels. We hypothesize that these dominant labels occupy large regions in the image space, and therefore represent good candidate labels for fooling most natural images. Note that these  dominant labels are automatically found by Algorithm \ref{alg:finding_universal_perturbations}, and are not imposed a priori in the computation of perturbations.

\textbf{Fine-tuning with universal perturbations.} We now examine the effect of fine-tuning the networks with perturbed images. We use the VGG-F architecture, and fine-tune the network based on a modified training set where universal perturbations are added to a fraction of (clean) training samples: for each training point, a universal perturbation is added with probability $0.5$, and the original sample is preserved with probability $0.5$.\footnote{In this fine-tuning experiment, we use a slightly modified notion of universal perturbations, where the \textit{direction} of the universal vector $v$ is fixed for all data points, while its \textit{magnitude} is adaptive. That is, for each data point $x$, we consider the perturbed point $x+\alpha v$, where $\alpha$ is the smallest coefficient that fools the classifier. We observed that this feedbacking strategy is less prone to overfitting than the strategy where the universal perturbation is simply added to all training points.} To account for the diversity of universal perturbations, we pre-compute a pool of $10$ different universal perturbations and add perturbations to the training samples randomly from this pool. The network is fine-tuned by performing $5$ extra epochs of training on the modified training set. To assess the effect of fine-tuning on the robustness of the network, we compute a new universal perturbation for the fine-tuned network (using Algorithm \ref{alg:finding_universal_perturbations}, with $p=\infty$ and $\xi = 10$), and report the fooling rate of the network. After $5$ extra epochs, the fooling rate on the validation set is $76.2\%$, which shows an improvement with respect to the original network ($93.7\%$, see Table \ref{tab:fooling_ratios}).\footnote{This fine-tuning procedure moreover led to a minor increase in the error rate on the validation set, which might be due to a slight overfitting of the perturbed data.} Despite this improvement, the fine-tuned network remains largely vulnerable to small universal perturbations. We therefore repeated the above procedure (i.e., computation of a pool of 10 universal perturbations for the fine-tuned network, fine-tuning of the new network based on the modified training set for $5$ extra epochs), and we obtained a new fooling ratio of $80.0\%$. In general, the repetition of this procedure for a fixed number of times did \textit{not} yield any improvement over the $76.2\%$ fooling ratio obtained after one step of fine-tuning. Hence, while fine-tuning the network leads to a mild improvement in the robustness, we observed that this simple solution does not fully immune against small universal perturbations.
\section{Explaining the vulnerability to universal perturbations}

\begin{figure}[t]
\centering
\includegraphics[width=0.4\textwidth]{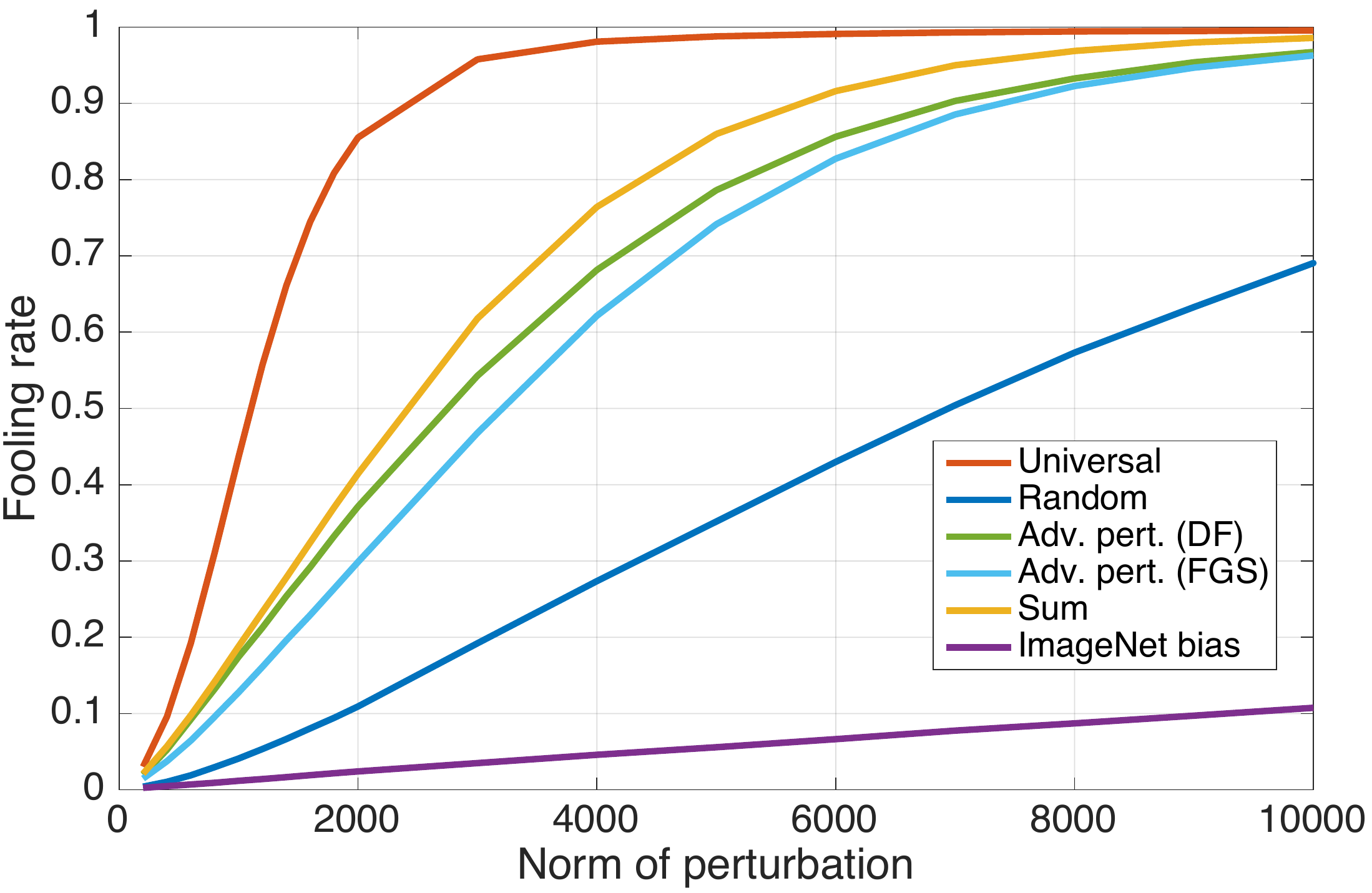}
\caption{\label{fig:curve_comparison} Comparison between fooling rates of different perturbations. Experiments performed on the CaffeNet architecture.}
\end{figure}

The goal of this section is to analyze and explain the high vulnerability of deep neural network classifiers to universal perturbations. To understand  the unique characteristics of universal perturbations,
we first compare such perturbations with other types of perturbations, namely
i) \textit{random} perturbation, ii) \textit{adversarial} perturbation computed for a randomly picked sample (computed using the DF and FGS methods respectively in \cite{moosavi2015deepfool} and \cite{goodfellow2014}),
iii) \textit{sum} of adversarial perturbations over $X$, and iv) mean of the images (or \textit{ImageNet bias}). For each perturbation, we depict a phase transition graph in Fig. \ref{fig:curve_comparison} showing the fooling rate on the validation set with respect to the $\ell_2$ norm of the perturbation. Different perturbation norms are  achieved by scaling accordingly each perturbation with a multiplicative factor to have the target norm. Note that the universal perturbation is computed for $\xi = 2000$, and also scaled accordingly.

Observe that the proposed universal perturbation quickly reaches very high fooling rates, even when the perturbation is constrained to be of small norm. For example, the universal perturbation computed using Algorithm \ref{alg:finding_universal_perturbations} achieves a fooling rate of $85\%$ when the $\ell_2$ norm is constrained to $\xi = 2000$, while other perturbations achieve much smaller ratios for comparable norms. In particular, random vectors sampled uniformly from the sphere of radius of $2000$ only fool $10\%$ of the validation set. The large difference between universal and random perturbations suggests that the universal perturbation exploits some \textit{geometric correlations} between different parts of the decision boundary of the classifier. In fact, if the orientations of the decision boundary in the neighborhood of different data points were completely uncorrelated (and independent of the distance to the decision boundary), the norm of the best universal perturbation would be comparable to that of a random perturbation. Note that the latter quantity is well understood (see \cite{nips2016_ours}), as the norm of the random perturbation required to fool a specific data point precisely behaves as $\Theta(\sqrt{d} \| r \|_2)$, where $d$ is the dimension of the input space, and $\| r \|_2$ is the distance between the data point and the decision boundary (or equivalently, the norm of the smallest adversarial perturbation). For the considered ImageNet classification task, this quantity is equal to $\sqrt{d} \| r \|_2 \approx 2 \times 10^4$, for most data points, which is at least one order of magnitude larger than the universal perturbation ($\xi = 2000$). This substantial difference between \textit{random} and \textit{universal} perturbations thereby suggests redundancies in the geometry of the decision boundaries that we now explore.

\begin{figure}[t]
\centering
\includegraphics[width=0.35\textwidth]{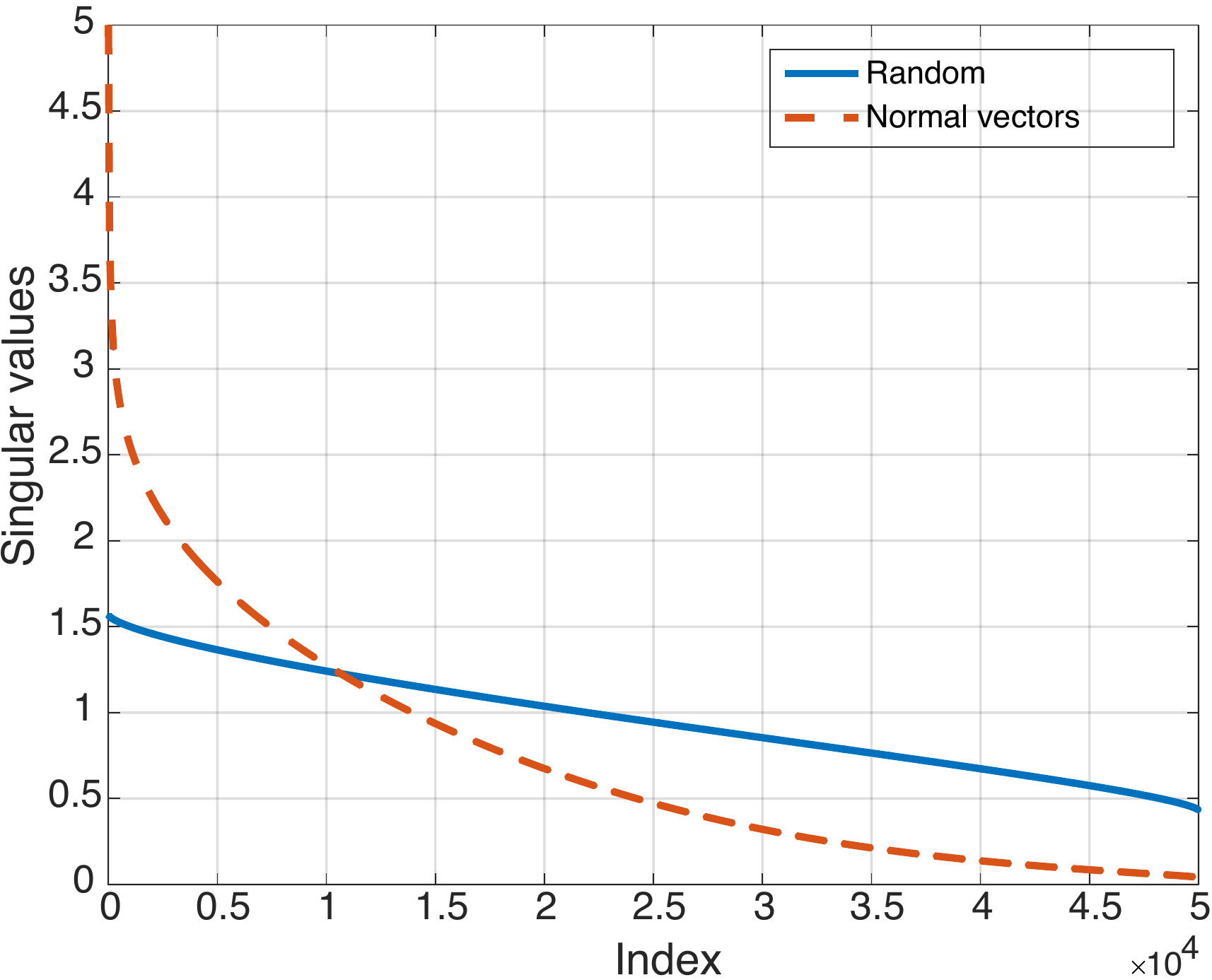}
\caption{\label{fig:singular_values} Singular values of matrix $N$ containing normal vectors to the decision decision boundary.}
\end{figure}

\begin{figure}[t]
\centering
\includegraphics[width=0.35\textwidth]{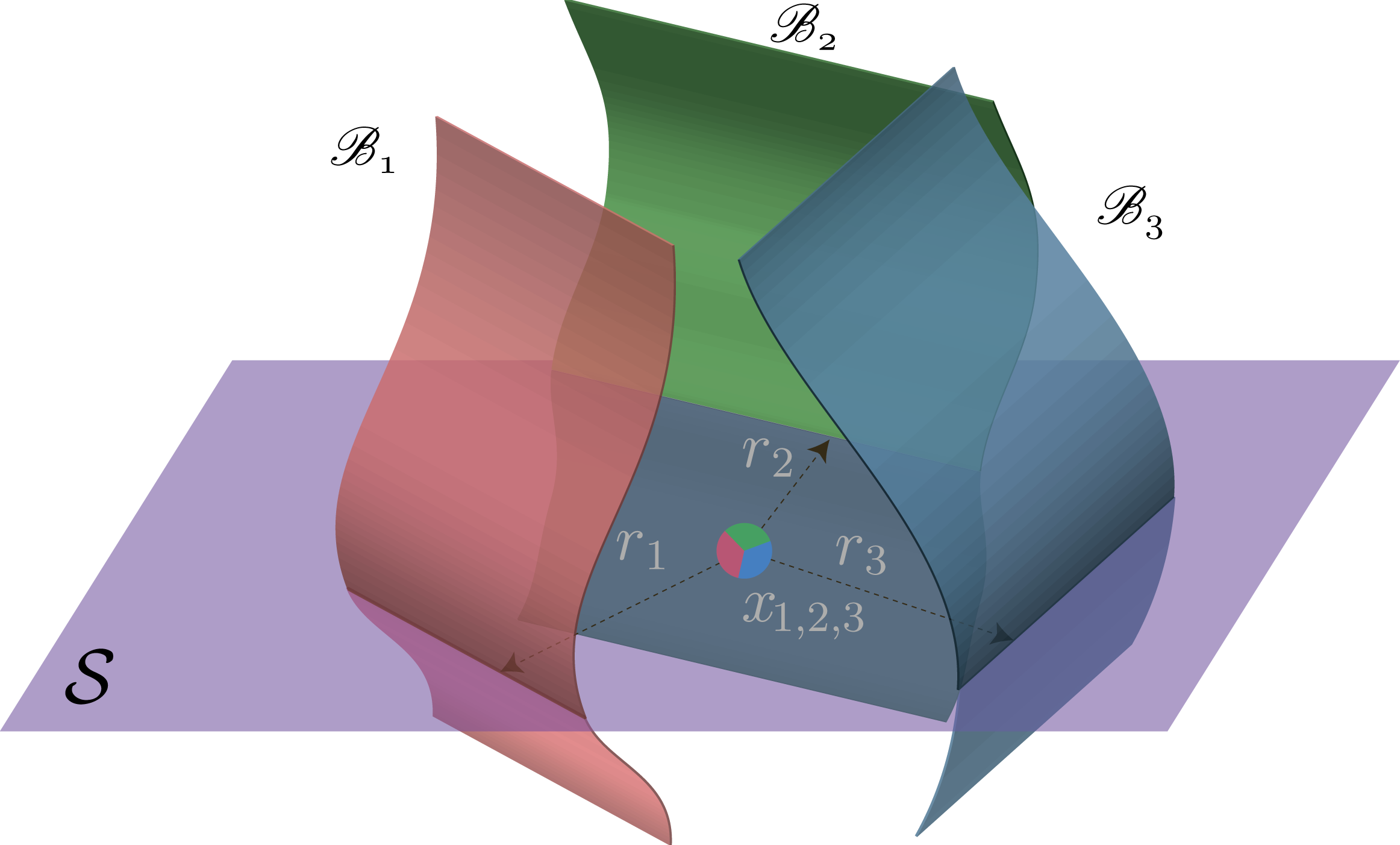}
\caption{\label{fig:subspace_illust} Illustration of the low dimensional subspace $\mathcal{S}$ containing normal vectors to the decision boundary in regions surrounding natural images. For the purpose of this illustration, we super-impose three data-points $\{x_i\}_{i=1}^3$, and the adversarial perturbations $\{r_i\}_{i=1}^3$ that send the respective datapoints to the decision boundary $\{\mathscr{B}_i\}_{i=1}^3$ are shown. Note that $\{r_i\}_{i=1}^3$ all live in the subspace $\mathcal{S}$.}
\end{figure}

For each image $x$ in the validation set, we compute the adversarial perturbation vector $r(x) = \arg\min_{r} \| r \|_2 \text{ s.t. } \hat{k} (x+r) \neq \hat{k} (x)$. It is easy to see that $r(x)$ is \textit{normal} to the decision boundary of the classifier (at $x + r(x)$). The vector $r(x)$ hence captures the local geometry 
 of the decision boundary in the region surrounding the data point $x$. To quantify the correlation between different regions of the decision boundary of the classifier, we define the matrix $$N = \left[ \frac{r(x_1)}{\| r(x_1) \|_2} \dots \frac{r(x_n)}{\| r(x_n) \|_2} \right]$$ of normal vectors to the decision boundary in the vicinity of $n$ data points in the validation set. For binary linear classifiers, the decision boundary is a hyperplane, and $N$ is of rank $1$, as all normal vectors are collinear. To capture more generally the  correlations in the decision boundary of complex classifiers, we compute the singular values of the matrix $N$. The singular values of the matrix $N$, computed for the CaffeNet architecture are shown in Fig. \ref{fig:singular_values}. We further show in the same figure the singular values obtained when the columns of $N$ are sampled uniformly at random from the unit sphere. Observe that, while the latter singular values have a slow decay, the singular values of $N$ decay quickly, which confirms the existence of large correlations and redundancies in the decision boundary of deep networks. More precisely, this suggests the existence of a subspace $\mathcal{S}$ of low dimension $d'$ (with $d' \ll d$), that contains most normal vectors to the decision boundary in regions surrounding natural images. We hypothesize that the existence of universal perturbations fooling most natural images is partly due to the existence of such a low-dimensional subspace that captures the correlations among different regions of the decision boundary. In fact, this subspace ``collects'' normals to the decision boundary in different regions, and perturbations belonging to this subspace are therefore likely to fool datapoints.
  To verify this hypothesis, we choose a \textit{random} vector of norm $\xi = 2000$ belonging to the subspace $\mathcal{S}$ spanned by the first $100$ singular vectors, and compute its fooling ratio on a different set of images (i.e., a set of images that have not been used to compute the SVD). Such a perturbation can fool nearly $38\%$ of these images, thereby showing that a \textit{random} direction in this well-sought subspace $\mathcal{S}$ significantly outperforms random perturbations (we recall that such perturbations can only fool $10\%$ of the data). Fig. \ref{fig:subspace_illust} illustrates the subspace $\mathcal{S}$ that captures the correlations in the decision boundary.
  It should further be noted that the existence of this low dimensional subspace explains the surprising generalization properties of universal perturbations obtained in Fig. \ref{fig:histogram_linf_numDatapointsX_val}, where one can build relatively generalizable universal perturbations with very few images. 

Unlike the above experiment, the proposed algorithm does \textit{not} choose a random vector in this subspace, but rather chooses a specific direction in order to maximize the overall fooling rate. This explains the  gap between the fooling rates obtained with the random vector strategy in $\mathcal{S}$ and Algorithm \ref{alg:finding_universal_perturbations}.

\section{Conclusions}
We showed the existence of small universal perturbations that can fool state-of-the-art classifiers on natural images. We proposed an iterative algorithm to generate universal perturbations, and highlighted several properties of such perturbations. In particular, we showed that universal perturbations generalize well across different classification models, resulting in doubly-universal perturbations (image-agnostic, network-agnostic). We further explained the existence of such perturbations with the correlation between different regions of the decision boundary. This provides insights on the geometry of the decision boundaries of deep neural networks, and contributes to a better understanding of such systems. A theoretical analysis of the geometric correlations between different parts of the decision boundary will be the subject of future research.


\subsection*{Acknowledgments}
We gratefully acknowledge the support of NVIDIA Corporation with the donation of the Tesla K40 GPU used for this research.
{\small
\bibliographystyle{ieee}
\bibliography{bibliography}
}

 \clearpage
 \onecolumn
 \appendix
 \section{Appendix}

\begin{figure}[h]
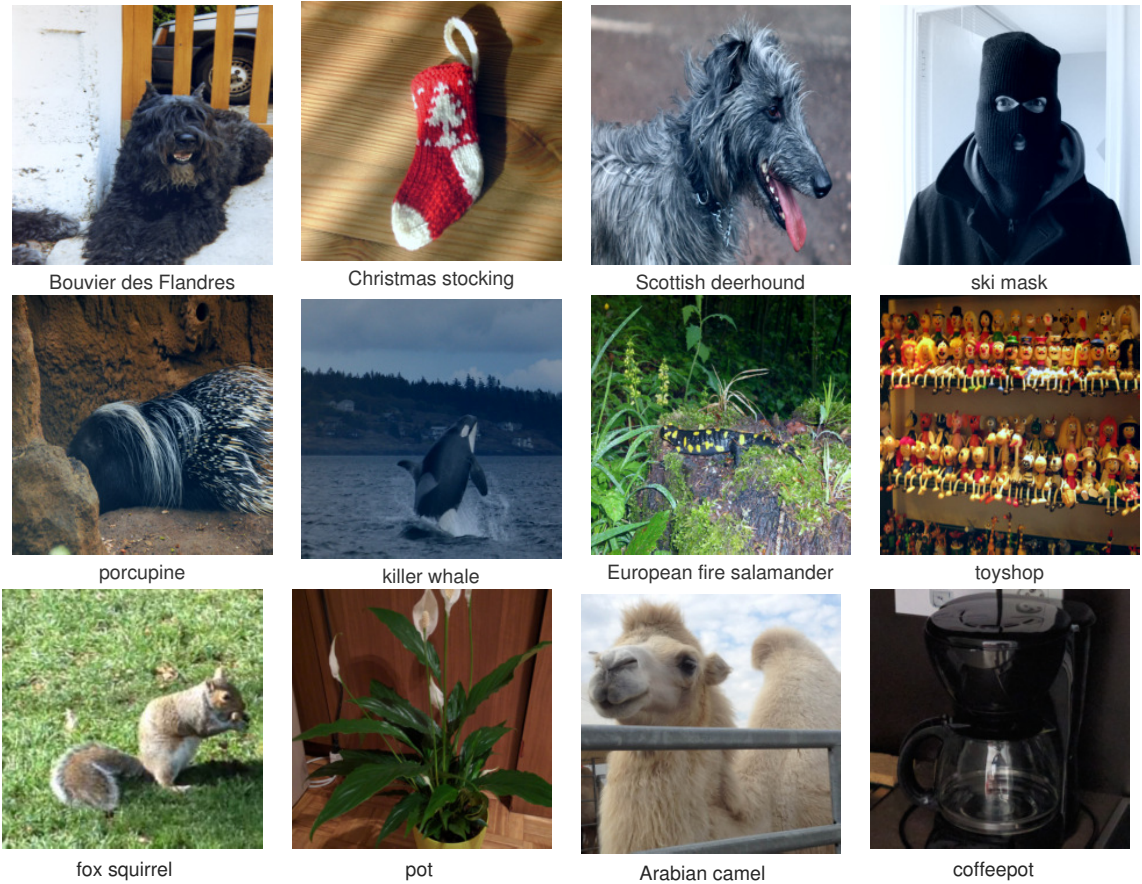

\centering
\newcounter{original}
\forloop{original}{1}{\value{original} < 9}{
\begin{subfigure}[t]{0.2\textwidth}
\includegraphics[width=\textwidth]{image_originalonly_\arabic{original}.pdf}
\end{subfigure}
~
}
\newcounter{natoriginal}
\forloop{natoriginal}{1}{\value{natoriginal} < 5}{
\begin{subfigure}[t]{0.2\textwidth}
\includegraphics[width=\textwidth]{natimg\arabic{natoriginal}_cropped_googlenet_orig.pdf}
\end{subfigure}
~
}
\caption{\label{fig:original_images} Original images. The first two rows are randomly chosen images from the validation set, and the last row of images are personal images taken from a mobile phone camera.}
\end{figure}

\begin{figure}[ht]
\centering
\includegraphics[width=1.0\textwidth]{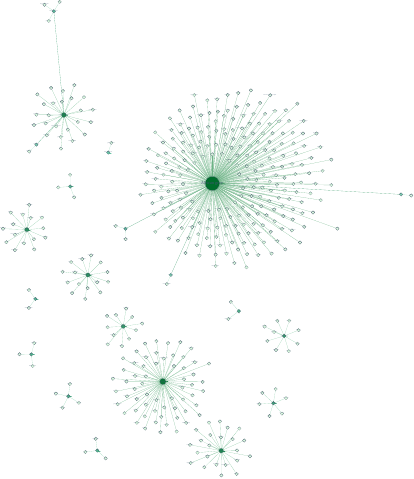}
\caption{\label{fig:graph_universal} Graph representing the relation between original and perturbed labels. Note that ``dominant labels'' appear systematically. Please zoom for readability. Isolated nodes are removed from this visualization for readability.}
\end{figure}

Fig. \ref{fig:original_images} shows the original images corresponding to the experiment in Fig. \ref{I-fig:perturbed_images}. Fig. \ref{fig:graph_universal} visualizes the graph showing relations between original and perturbed labels (see Section \ref{I-sec:experiments} for more details).

\label{app:app1}

\end{document}